\documentclass[letterpaper, 10 pt, conference]{ieeeconf}  

\IEEEoverridecommandlockouts                              

\usepackage{graphics} 
\usepackage{graphicx}
\usepackage{amsmath} 
\usepackage{amssymb}  
\usepackage{mathtools}
\usepackage{hyperref}
\usepackage[ruled,vlined]{algorithm2e}

\usepackage[symbol]{footmisc}

\title{\LARGE \bf
Rapid Pose Label Generation through \protect\\ Sparse Representation of Unknown Objects
}

\author{Rohan P. Singh$^{1,2}$, Mehdi Benallegue$^{1}$, Yusuke Yoshiyasu$^{3}$, Fumio Kanehiro$^{1,2}$
\thanks{$^{1}$Rohan P. Singh, Mehdi Benallegue, Yusuke Yoshiyasu and Fumio Kanehiro are with CNRS-AIST JRL (Joint Robotics Laboratory) IRL, National Institute of Advanced Industrial Science and Technology (AIST), Japan
        {\tt\small rohan-singh@aist.go.jp}}%
\thanks{$^{2}$Rohan P. Singh and Fumio Kanehiro are with University of Tsukuba, Ibaraki, Japan}%
}

\begin{document}

\maketitle
\thispagestyle{empty}
\pagestyle{empty}

\begin{abstract}

Deep Convolutional Neural Networks (CNNs) have been successfully deployed on robots for 6-DoF object pose estimation through visual perception. However, obtaining labeled data on a scale required for the supervised training of CNNs is a difficult task - exacerbated if the object is novel and a 3D model is unavailable. To this end, this work presents an approach for rapidly generating real-world, pose-annotated RGB-D data for unknown objects. Our method not only circumvents the need for a prior 3D object model (textured or otherwise) but also bypasses complicated setups of fiducial markers, turntables, and sensors. With the help of a human user, we first source minimalistic labelings of an ordered set of arbitrarily chosen keypoints over a set of RGB-D videos. Then, by solving an optimization problem, we combine these labels under a world frame to recover a sparse, keypoint-based representation of the object. The sparse representation leads to the development of a dense model and the pose labels for each image frame in the set of scenes. We show that the sparse model can also be efficiently used for scaling to a large number of new scenes. We demonstrate the practicality of the generated labeled dataset by training a pipeline for 6-DoF object pose estimation and a pixel-wise segmentation network.

\end{abstract}

\section{INTRODUCTION}

Object manipulation tasks performed by autonomous robots require the robot to recognize and locate the object in the 3D space, typically through visual perception. Precise grasping tasks necessitate an estimation of the full 6-DoF pose of task-relevant objects from the input visual data. In the past few years, powerful CNNs have been successfully employed for this purpose enabling object pose estimation under difficult circumstances \cite{pavlakos17object3d, xiang2018posecnn}. However, the performance of most CNN based state-of-the-art methods is dependent on the availability of large sets of real-world data --- labeled with the ground-truth pose for each object instance --- to supervise the training of the involved neural networks. This training dataset may easily consist of tens of thousands or more labeled data points per object in varying backgrounds and environments \cite{xiang2018posecnn, zeng2017multi:zengapc, rad2017bb8:bb8}. In such cases, manual labeling of raw samples becomes unreasonable and impractical. Other pose estimation methods that claim to use artificially generated synthetic data are either still partially dependent on real data for fine-tuning or require very high quality textured 3D object models \cite{tremblay2018corl:dope}. Nevertheless, real-world annotated data is indispensable for a meaningful and comprehensive evaluation of any pose estimator.


Techniques for (semi-) automating the process of training data generation has gained research interest in recent years \cite{rennie2016dataset, wong2017segicp, hodan2017t:tless, hinterstoisser2012model:linemod, suzui2019toward}. However, their reliance on additional hardware such as multi-sensor rigs, turntables with markers, or motion capture systems limits their application in arbitrary environments. Besides, hardware setup and sensor calibration may in itself be time-consuming. LabelFusion \cite{marion2018label} is a recent but popular method that allows raw data capture through a hand-held RGB-D sensor. Yet, its dependence on the availability of pre-built object meshes restricts operation on a wider object set. EasyLabel \cite{suchi2019easylabel} attempts to overcome the dependence on pre-built models. However, it cannot produce data on a scale that is necessary for training deep networks, as the method works on individual snapshots of the scene and there is no propagation of labels. 

We address these problems by proposing a technique to produce large amounts of pose-annotated, real-world data for rigid objects that uses minimal human input and does not require any previously built 3D shape or texture model of the object. Our method also simplifies the capture of raw (unlabeled, unprocessed RGB-D) data due to independence from turntables, marker fields, motion capture setups, and manipulator arms. The key idea is to use sparse annotations provided manually and systematically by a human user over multiple scenes, and combine them under geometrical constraints to produce the labels. We define ``pose labels" as the 2D bounding-box labels, 2D keypoint labels and the pixel-wise mask labels (although other types of labels can be deduced too). Hence, our method can effectively be used for generating large training datasets -- for 2D/3D object detection, 2D/3D keypoint estimation, 6-DoF object pose estimation, and semantic segmentation -- for novel objects in real scenarios outside of popular datasets such as LineMOD\cite{hinterstoisser2012model:linemod} and T-LESS\cite{hodan2017t:tless}.

We demonstrate the effectiveness of our method by generating keypoint labels, pixel-wise mask labels, and bounding-box labels for more than 150,000 RGB images of 11 unique objects (7 in-house + 4 YCB-Video \cite{xiang2018posecnn} objects) in a total of 80 (single and multi-object) scenes --- in only a few minutes of manual annotation for each object. We evaluate the accuracy of the generated labels and the sparse model using ground-truth CAD models. Subsequently, we train and evaluate (1) a keypoint-based 6-DoF pose estimation pipeline and (2) an object segmentation CNN using the resulting labeled dataset. Our proposed tool with the graphical user-interface are published as a public GitHub repository \footnote[2]{\url{https://github.com/rohanpsingh/RapidPoseLabels}}.

\section{RELATED WORK}

Various works on developing open-source 6-DoF pose labeled datasets have adopted different approaches for automating the process of labeling RGB-D data \cite{hinterstoisser2012model:linemod, hodan2017t:tless, rennie2016dataset, hua2016scenenn, dai2017scannet}. The popular Rutger's dataset \cite{rennie2016dataset} was generated by mounting a Kinect sensor to the end joint of a Motoman robot arm. The annotation was done by a human to align the 3D model of the objects in the corresponding RGBD point cloud scene. This approach does produce high quality ground-truth data albeit at the cost of laborious manual involvement. The T-LESS dataset \cite{hodan2017t:tless}, containing about 50K RGB-D images of 30 textureless objects, was developed with an arguably complicated setup involving a turntable with marker-field and a triplet of sensors attached to a jig. The method for dataset generation described in \cite{zeng2017multi:zengapc} works through background subtraction, first recording data in a scene without the object and then with the object. This can produce pixel-wise segmentation labels even for model-less objects, but cannot be used for 6-DoF pose labels. SegICP \cite{wong2017segicp} used a labeled set of 7500 images in indoor scenes, produced with the help of a motion capture (MoCap) based automated annotation system, which inevitably limits the types of environments in which data is acquired. An interesting method of collecting data through the robot in a life-long self-supervised learning fashion was presented in \cite{deng2019self}. Here the robot learns pose estimation and simultaneously generates new data for improving pose estimation by itself, although in a very limited environment.

The complications involved in acquisition of real-world training data have inspired methods such as \cite{tremblay2018corl:dope, tobin2017domain} which are trained only on synthetic or photo-realistic simulated data. Although demonstrating very promising results, they present an additional requirement of the availability of a very high quality textured 3D model, which may be hard to obtain in itself without dedicated special hardware.

Marion et. al. presented LabelFusion \cite{marion2018label}  -- a method for generating ground-truth labels for RGB-D data minimizing the human effort involved. They perform a dense 3D reconstruction of the scene using ElasticFusion, and then ask the user to manually annotate 3 points in the 3D scene space to initiate an ICP-based registration to align a previously built 3D model to the scene point cloud. For sidestepping the need for a prior object model, Suchi et. al. proposed EasyLabel \cite{suchi2019easylabel} wherein scenes are incrementally built and depth changes are exploited to obtain object masks. This however limits the scale at which data can be generated and so, their method is presented primarily for evaluation purposes -- as opposed to training of deep CNNs.

Our method outperforms the current state-of-the-art in regard to the above-mentioned problems. It is capable of labeling large scales of datasets, suitable for supervised training of deep networks and, as stated earlier, it is not dependent on the availability of a prior 3D object model.

\begin{figure*}
  \includegraphics[width=\textwidth]{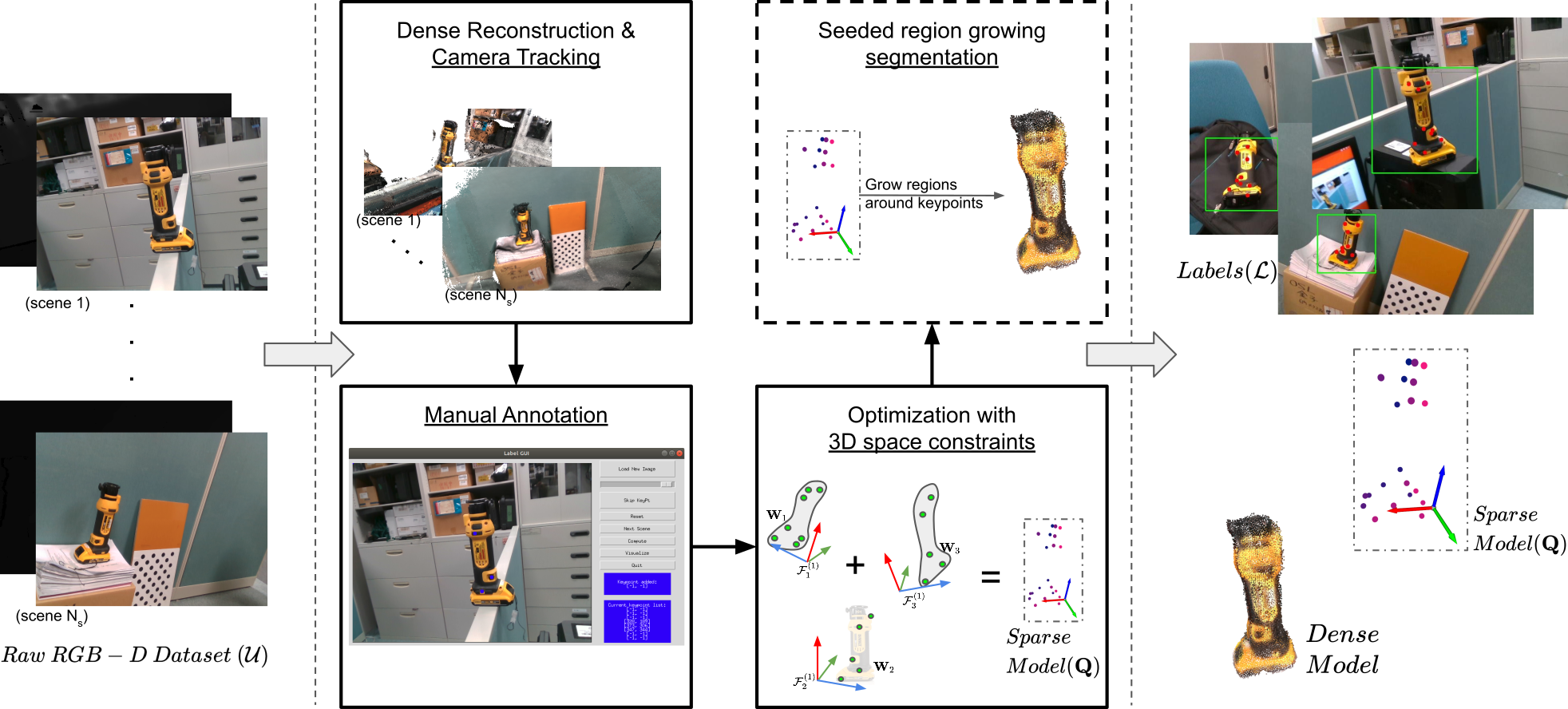}
  \caption{\textbf{Proposed system for generating pose labels.} The input to the system is a set of RGB-D image sequences and a limited amount of manual labels. The output consists of (1) labels for each frame in the raw dataset, (2) a sparse, keypoint-based representation and (3) a dense model of the object. Existing techniques are used for dense reconstruction while a user-friendly GUI is implemented for providing the manual annotations. An optimization problem is solved on the manual annotations to recover a sparse model and finally, a dense model is built-up from the sparse model using a region-growing segmentation algorithm.}
  \label{figure:main_approach}
\end{figure*}

\section{APPROACH}

We consider the case of generating training data for a single object-of-interest. The human user collects a set of $N_s$ RGB-D videos (alternatively referred to as \textit{scenes}) by moving a hand-held RGB-D sensor around the relevant object --- placed in varying orientations, positions, backgrounds, and lighting conditions. This set of unlabeled videos or scenes is denoted by $\mathcal{U} = \{\mathcal{U}_1, \mathcal{U}_2, \dots, \mathcal{U}_{N_s}\}$. The duration and frame rate of each video $\mathcal{U}_s$ can be variable (2-5 minutes and 30fps in our experiments). Mathematically, a scene consisting of ${t_{s}}$ frames is defined as $\mathcal{U}_s = \{ (\mathbf{I}_s^{(t)}, \mathbf{D}_s^{(t)})_{t=1}^{t_{s}} \}$, where $\mathbf{I}_s^{(t)}$ and $\mathbf{D}_s^{(t)}$ are the RGB and depth images respectively at time instance $t \in \{1 ,\dots, {t_{s}}\}$ in the scene $s \in \{1 ,\dots, {N_s}\}$. Our final objective is to generate the labeled dataset $\mathcal{L} = \{\mathcal{L}_1, \mathcal{L}_2, \dots, \mathcal{L}_{N_s}\}$, i.e. associate each frame in each scene with a pose label.

\textbf{System Overview.} Figure \ref{figure:main_approach} illustrates the overview of the process. Given $\mathcal{U}$, our system initiates by performing dense scene reconstructions using a third-party software to obtain scene meshes and recover camera trajectories (Section: \ref{subsec:dense_recon}). Next, we obtain the manual labels with the help of our GUI tool (Section \ref{subsec:manual_annot}). The key idea in this step is to ask the human annotator to label an ordered set of $N_k$ arbitrarily chosen landmark points on the object (called the \textit{keypoints set}) in each scene. The user needs to sequentially mark only a subset of the keypoints on any randomly selected RGB image in each collected scene. This produces fragments of the sparse keypoint-based object representation each defined in the respective scene's frame of reference. If there is enough overlap between the fragments, we can recover simultaneously (1) the full, 3D keypoint-based sparse model $\mathbf{Q} \in \mathbb{R}^{3\times N_k}$ and (2) the relative transformations between the scenes, explaining the annotations provided by the user. We achieve this by formulating and solving an optimization problem on the sparse, user-annotated keypoint configurations constrained on 3D space rigidity (Section \ref{subsec:optim_step}). The resulting output of a successful optimization is (optionally, if mask labels are required) used to build a dense model of the object by segmenting out the points corresponding to the object of interest from all scene meshes and combining them together (Section \ref{subsec:dense_model}). Finally, the produced sparse and dense models can be projected back to all the 2D image planes in $\mathcal{U}$ to obtain the desired type of labels (Section \ref{subsec:label_generation}). 

\subsection{Dense Scene Reconstruction} \label{subsec:dense_recon}
Each scene $\mathcal{U}_s$ is a sequence of image pairs where the camera trajectory through time is unknown. It is valuable to obtain this trajectory as it can be used to automatically propagate labels through all instances in the scene. We rely on an existing technique which provides camera pose tracking along with a dense 3D reconstruction, to avoid dependence on fiducial markers or robotic manipulators. With this, the dataset $\mathcal{U}$ is now altered to become $\mathcal{U} = \{\{(\mathbf{I}_s^{(t)}, \mathbf{D}_s^{(t)}, \mathbf{C}_s^{(t)})_{t=1}^{t_{s}}\}_{s=1}^{N_s}\}$, where $\mathbf{C}_s^{(t)} \in \mathbb{R}^{4\times4}$ is the homogeneous transformation matrix giving the camera pose $\mathcal{F}_{s}^{(t)}$ at instance $t$ in the scene $s$ relative to the initial camera pose $\mathcal{F}_{s}^{(1)}$ of the same scene. In our experiments, we use the open-source implementation of ElasticFusion \cite{whelan2016elasticfusion} for this step (as in \cite{marion2018label}).
It is worthwhile to mention here that while recording the videos, an individual scene $\mathcal{U}_s$ does not necessarily need to consist of a full scan of the object from all views, which may be difficult to obtain. Our method combines different object views from all scenes in $\mathcal{U}$ to produce final output.

\subsection{Manual Annotation} \label{subsec:manual_annot}
Keeping in line with our desire to reduce the involved human effort and time to as low as possible, our system sources minimal annotations from the human user. The user first chooses a set of arbitrary but well-distributed, ordered, and uniquely identifiable landmark points (called \textit{keypoints}) on the physical object. Then, with the help of our user-interface, the user sequentially labels the location of these keypoints in each scene. The labeling is done on the RGB image. As all faces of the object may not be visible in a scene, labeling is done only for the visible subset of the keypoints on the RGB image. The non-visible keypoints are skipped. Moreover, the annotations can be distributed over multiple images of the same scene.

For each user-annotated keypoint $k' \in \{1,\dots,N_k\}$ on the RGB image $\mathbf{I}_{s'}^{(t')}$ in scene $s' \in \{1 ,\dots, {N_s}\}$ and at time instance $t' \in \{1,\dots, {t_{s'}}\}$, using camera intrinsics of the RGB sensor and the depth image $\mathbf{D}_{s'}^{(t')}$, we obtain the 3D point position $\prescript{}{k'}{\mathbf{d}_{s'}^{(t')}} = [t_x, t_y, t_z]^T$ of the labeled pixel. Then, we transform this point $\prescript{}{k'}{\mathbf{d}_{s'}^{(t')}}$ to the coordinate frame $\mathcal{F}_{s'}^{(1)}$ (i.e. in the camera frame at time $t=1$ in scene $s=s'$):

\begin{equation} \label{eq:tf}
\begin{bmatrix}
\prescript{}{k'}{\mathbf{d}_{s'}^{(1)}} \\[6pt]
1
\end{bmatrix} = \mathbf{C}_{s'}^{(t')} \cdot
\begin{bmatrix}
\prescript{}{k'}{\mathbf{d}_{s'}^{(t')}} \\[6pt]
1
\end{bmatrix}.
\end{equation}

For each scene $s$, this gives us the matrix $\mathbf{W}_{s} \in \mathbb{R}^{3 \times N_k}$, where the columns hold the 3D position $\prescript{}{k}{\mathbf{d}_{s}^{(1)}}$ of the keypoint $k$ if it was manually annotated and $\mathbf{0}^{3}$ otherwise. Doing so for all scenes, we obtain $\mathcal{W} = \{\mathbf{W}_1, \mathbf{W}_2, \dots, \mathbf{W}_{N_s}\}$.
    
\textbf{Selection of Keypoints.} Although the selection of keypoints on the object is arbitrary, there are some constraints on the manual annotation which must be kept in mind by the user while choosing them. To ensure existence of a unique solution to the optimization problem, the marked keypoints should rigidly connect all the scenes together, namely by avoiding the system to suffer underdetermination in the produced constraints for the scene relative poses. For example, for two scenes, mutual sharing of 3 or more non-collinear keypoints rigidly ``ties" them. Hence, we recommend choosing points that are more susceptible to be shared in multiple scenes, such as on the edge shared by two faces of an object.

\subsection{Optimization} \label{subsec:optim_step}
Thus far, we have localized subsets or fragments of the object's keypoint model $\mathbf{Q}$ in each scene of $\mathcal{U}$. The subsets do not exist in one common frame of reference. Instead, the fragment $\mathbf{W}_{s}$ is defined in the coordinate frame $\mathcal{F}_{s}^{(1)}$ and the relative transformations between scenes are yet unknown. In this step we find the relative transformations and combine the fragments in a common space to recover $\mathbf{Q}$. Let us represent the set of relative transformations by $\mathcal{T} = \{{\mathbf{T}_{1}}, {\mathbf{T}_2}, \dots, {\mathbf{T}_{N_s}}\}$ where $\mathbf{T}_{s} = \begin{bmatrix} \mathbf{q}_s \quad \mathbf{t}_s \end{bmatrix}$ is the rigid transformation of the local frame $\mathcal{F}_{s}^{(1)}$ with respect to the world frame $\mathcal{F}_{w}$. $\mathbf{q}_s \in \mathbb{R}^4$ represents the rotation quaternion and $\mathbf{t}_s \in \mathbb{R}^3$ is the 3D position of the camera center. We set $\mathcal{F}_{w} = \mathcal{F}_{1}^{(1)}$ (i.e. origin of scene $1$). We formulate the following nonlinear optimization problem:

\begin{align}
\mathbf{Q}\text{*},\mathcal{T}\text{*} & = \underset{\mathbf{Q},\mathcal{T}}{\text{argmin}}. 
\|  \mathbf{S} \cdot f(\mathcal{T}, \mathbf{Q}) - \mathbf{W} \| ^2, \nonumber \\
& \textrm{s.t.} \quad \mathbf{q}_s^T\cdot\mathbf{q}_s=1, \nonumber \\
& \forall s \in \{0 ,\dots, N_s-1\}.  \label{eq:opt_eq}
\end{align}
\noindent
where $\mathbf{W}$ is the concatenation of all non-zero points in $\mathcal{W}$ and the function $f(\cdot)$ successively applies each $\mathbf{T}_{s} \in \mathcal{T}$ to $\mathbf{Q}$ and returns the concatenated vector. The selection matrix $\mathbf{S}$ selects from the vector $f(\mathcal{T}, \mathbf{Q})$ only those keypoints whose reference is available in $\mathcal{W}$, that is, the keypoints which were manually annotated. The solution is given by $\mathbf{Q}\text{*}$ - representing the optimized keypoint representation of the object model defined in the frame $\mathcal{F}_{w}$ - and the set $\mathcal{T}\text{*}$ - representing the optimized relative scene transformations.

\begin{figure}[b]
  \includegraphics[width=\linewidth]{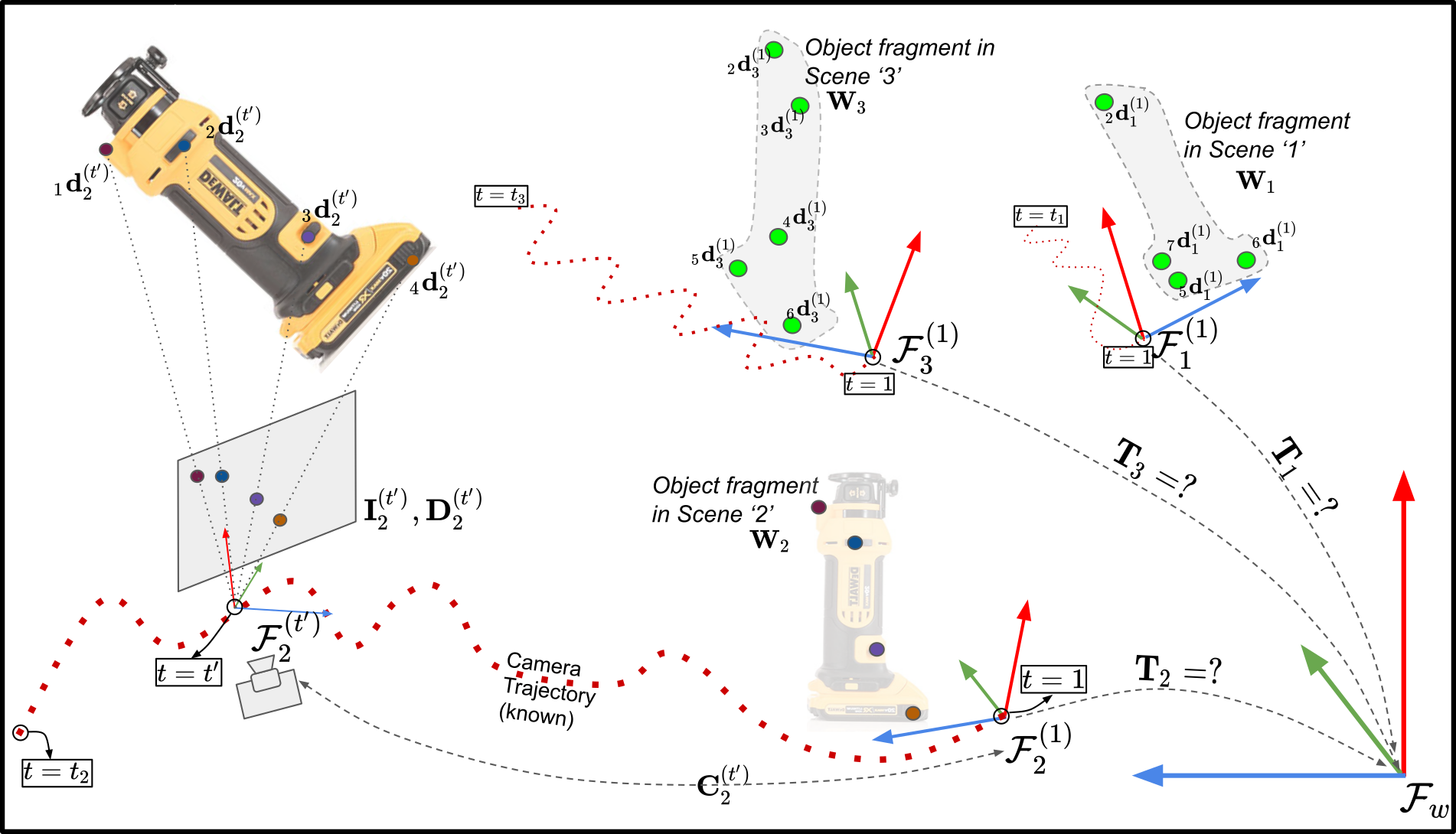}
  \caption{\textbf{Problem description.} An example dataset with 3 scenes `1', `2' and `3' with trajectory lengths $t_1$, $t_2$ and $t_3$ respectively are shown. The left-side shows 4 keypoints annotated on an image of Scene `2' at time $t=t'$, projected to get the 3D positions. This is done for each scene (with some amount of mutually shared keypoints). The 3D point fragments are then transformed to the origin of their respective scene trajectories, giving us $\mathbf{W}_1$, $\mathbf{W}_2$ and $\mathbf{W}_3$, on the right. The intention of the optimization, essentially, is to find $\mathbf{T}_1$, $\mathbf{T}_2$, $\mathbf{T}_3$ such that all fragments can be represented in world frame $\mathcal{F}_w$. Keypoint overlapping ensures existence of a unique solution(eg. keypoints 2,3,4 are annotated in scene `1' and `2' both).}
  \label{figure:optimization}
\end{figure}

Intuitively, solving Eq. \ref{eq:opt_eq} is equivalent to bringing together the subsets in $\mathcal{W}$, defined in their local frames, into one common world frame of reference (as illustrated in Figure \ref{figure:optimization}). This is possible because the user has inputted mutually overlapping annotations, and the solution to Eq. \ref{eq:opt_eq} provides a consistent explanation of all the observations.

We used the scipy.optimize library in Python with the Sequential Least Squares Programming solver \cite{kraft1988software} to minimize Eq. \ref{eq:opt_eq}. We initialized the iterations by setting all elements of $f(\mathcal{T}, \mathbf{Q})$ to 0 except for the real part of the rotation quaternion which is set to 1.

\subsection{Segmentation of Dense Object Model} \label{subsec:dense_model}
In this step, we use the sparse object model and the scene transformations produced above to segment out the 3D points corresponding to the object from each of the dense scenes, thereby producing a dense model of the object. A dense object model is necessary to obtain pixel-wise mask labels and is also useful for planning robotic grasp poses. We modify Point Cloud Library's (PCL) region-growing-segmentation algorithm such that the individual points of the sparse model act as seeds and the regions are grown and spread outwards from the seeds. All segmented regions from each scene are then combined using the relative transformations to create the dense model. The output may occasionally have holes in the geometry or contain points from the background scenes which require manual cropping, yet, as our experiments indicate, it remains a practical approximation for the object shape. 


\subsection{Generation of Pose Labels} \label{subsec:label_generation}
Having obtained the object sparse model, the dense model (both defined in $\mathcal{F}_{w}$) and the set $\mathcal{T}$ of scene transformations w.r.t. $\mathcal{F}_{w}$, the labeled dataset $\mathcal{L} = \{\mathcal{L}_1, \mathcal{L}_2,\dots, \mathcal{L}_{N_s}\}$ can be generated through back projection to the RGB image planes. For the purpose of training the 6-DoF pose estimation pipeline adopted by us \cite{pavlakos17object3d}, we define $\mathcal{L} = \{\{(\mathbf{I}_{s}^{(t)}, \textbf{L}_{s}^{(t)}, \textbf{b}_{s}^{(t)})_{t=1}^{t_{s}}\}_{s=1}^{N_s}\}$ where $\textbf{L}_{s}^{(t)} \in \mathbb{R}^{2\times{N_k}}$ is the 2D keypoint annotation in the image $\mathbf{I}_{s}^{(t)}$ for the $N_k$ keypoints chosen on the object and $\textbf{b}_{s}^{(t)} \in \mathbb{R}^{3}$ represents the $x,y$ pixel coordinates of the center and side length $h$ of an upright bounding-box square around the object in the image.

Computation of the label for the $k^{th}$ keypoint at time instance $t$ of the scene $s$, $\prescript{}{k}{\textbf{l}}_{s}^{(t)}$, can be done simply by transforming the point $\mathbf{Q}_{*,k}$ to the camera frame $\mathcal{F}_{s}^{(t)}$ to obtain the 3D point $\prescript{}{k}{\mathbf{d}_{s}^{(t)}} = [t_x, t_y, t_z]^T$, following Eq. \ref{eq:tf_inverse}. And subsequently, projecting it onto the 2D image plane to get $\prescript{}{k}{\textbf{l}}_{s}^{(t)} = [u_x, u_y]^T$. \par

\begin{align}
\begin{bmatrix}
\prescript{}{k}{\mathbf{d}_{s}^{(t)}} \\[6pt] 
1
\end{bmatrix}
 & = (\mathbf{C}_{s}^{(t)})^{-1} \cdot
 \begin{bmatrix}
{\mathbf{T}^{-1}_{s}} \circ {\mathbf{Q}_{*,k}} \\[6pt]
1
\end{bmatrix}. \label{eq:tf_inverse}
\end{align}

The per-pixel mask label can also be obtained similarly once the dense model has been generated - by transformation and back projection of each point of the 3D dense model onto the 2D RGB images. 

The bounding-box label $\textbf{b}_{s}^{(t)}$ can be obtained by finding the up-right bounding rectangle of the mask points or taking the bounding-box of all points in $\textbf{L}_{s}^{(t)}$ and scaling up by a factor of $1.5$ to avoid cropping out of the object itself.

\subsection{Using the Sparse Model for New Scenes}
Application of deep-learning approaches for practical purposes often requires fine tuning of the trained model with labeled data collected in the real environment of operation \cite{zeng2017multi:zengapc}. So, the user must be capable of quickly labeling a new RGB-D scene that is outside of the initial dataset $\mathcal{U}$. This is achieved as follows:

Once a model $\mathbf{Q}$ of an object has been generated from $\mathcal{U}$, for each new RGB-D video, the user must uniquely mark any subset of $\mathbf{Q}$ (at least 3 points) on a randomly selected RGB image. Then, we obtain the 3D positions of the points using depth data, and apply Procrustes Analysis (using Horn's solution \cite{horn1987closed}) to compute the rigid transformation between the set of marked points and their correspondences in $\mathbf{Q}$. We can generate the labels again through back projection of the dense model according to the computed transformation. As this process involves only a few seconds of manual annotation, scalability to a large number of scenes is convenient.

\begin{figure}[t]
  \includegraphics[width=\linewidth]{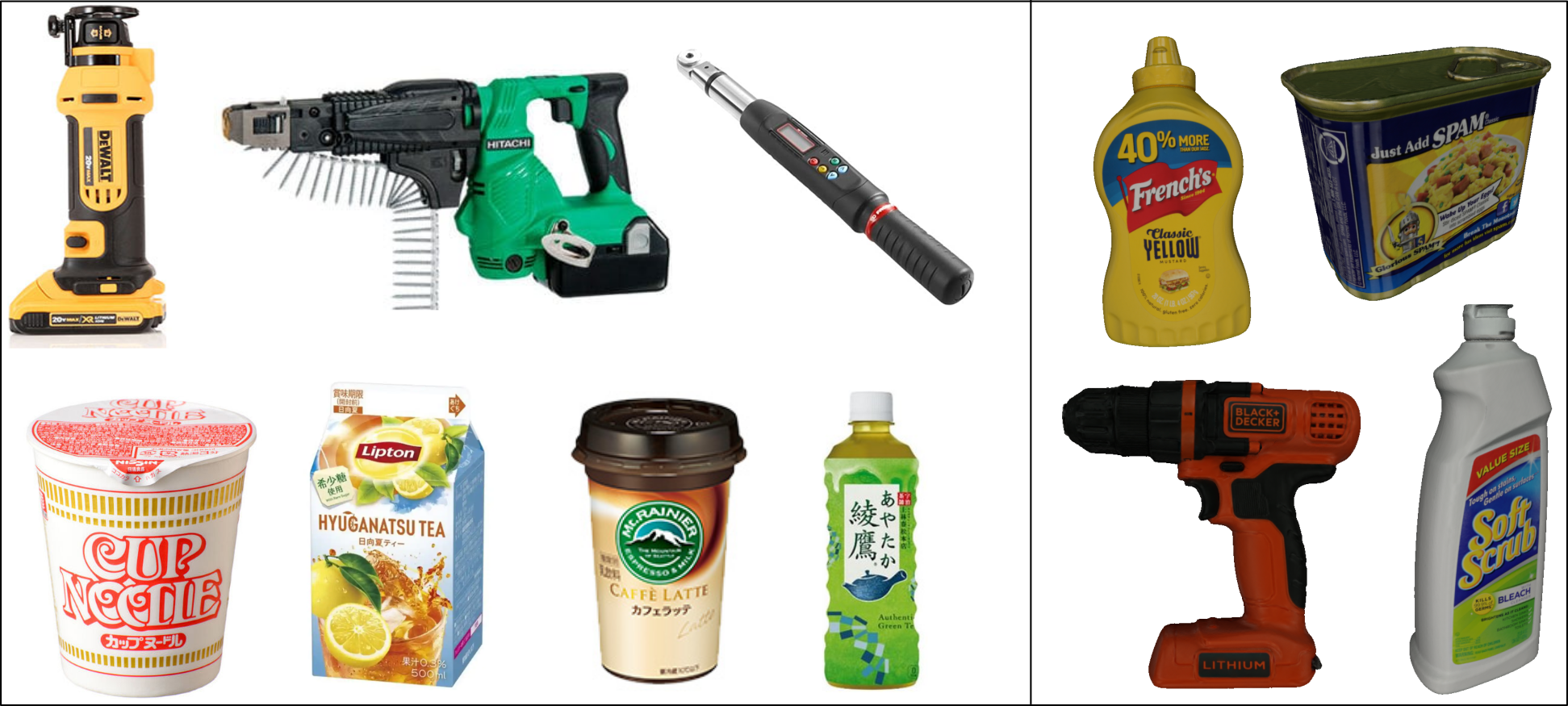}
  \caption{\textbf{Selected objects for experiments.} (a) In-house objects on the left and (b) objects selected from YCB-Video dataset on the right.}
  \label{figure:objects}
\end{figure}

\section{EXPERIMENTS}
We employ our proposed system to generate the sparse models, dense models and labeled datasets for two separate sets of objects for quantitative analysis (refer to Figure \ref{figure:objects}). The first set consists of 4 objects from the YCB-Video dataset \cite{xiang2018posecnn} which already provides us with multi-object RGB-D video scenes and the object CAD models. The second is a set of 7 objects for which we manually collected the RGB-D scenes with an Intel RealSense D435 at $640\times480$ resolution. The CAD models of the 7 in-house objects were either sourced from the manufacturer or drawn manually prior to the experiments. We name these objects models as GT-CAD. Nevertheless, the CAD models for all objects and experiments were used only for evaluation purposes, as a key advantage of our system is its independence from prior object models.

From the YCB-Video dataset, we selected 25 (20 single-object + 5 multi-object) scenes, with each of the 4 chosen objects appearing in 9 unique scenes. For the in-house objects, we recorded 55 (43 single-object + 12 multi-object) RGB-D scenes using the Intel RealSense D435 sensor. Each object's dataset $\mathcal{U}$ was carved out of these 55 + 25 scenes.

\subsection{3D Sparse Model and Label Generation} 
To evaluate the accuracy of the predicted sparse model, we align it to the corresponding GT-CAD model using ICP (with manual initialization) and find the closest points on the CAD model from each of the points in the sparse model. The set of closest points acts as the ground-truth for the sparse model (denoted by GT-SPARSE). We report the mean Euclidean distance between the corresponding points as an estimate of the sparse-model's accuracy. Next, as the primary motivation of our system is to enable automated generation of training data, we also evaluate the accuracy of the generated labels (keypoint + pixel-wise labels). To get the ground-truth for the keypoint and mask labels, we align (manual initialization followed by ICP fine tuning) the GT-CAD to the scene reconstructions, and project back GT-SPARSE and GT-CAD respectively to the RGB images using the camera trajectories (estimated by ElasticFusion or provided in YCB-Video dataset). Note that this approach is similar to the method of label generation in \cite{pavlakos17object3d, marion2018label}. 

Errors in keypoints labels are computed as the mean 2D distance between the predicted pixel coordinates and the ground-truth. For the pixel-wise mask labels, we use the popular Intersection-over-Union (IoU) metric. Table \ref{tab:table_model_errors} lists the results of this quantitative analysis for each object along with the number of scenes and number of keypoints chosen for this set of experiments.

\begin{figure}[b]
  \includegraphics[width=\linewidth]{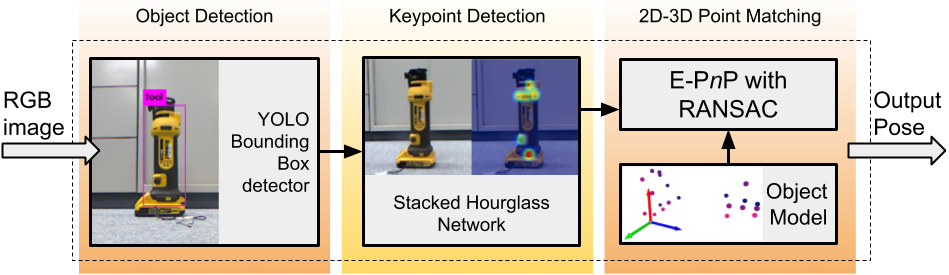}
  \caption{\textbf{Block diagram of adopted pose estimation pipeline.} Computed by solving a P\textit{n}P problem on predicted 2D keypoints and their 3D correspondences.}
  \label{figure:pose_estimation}
\end{figure}

The errors in the mask labels also give an indication of the accuracy of the dense model. As explained earlier, the dense model requires manual cropping (~2-3min), but our system reduces the overall time in creating labeled datasets by several orders of magnitude as compared to the completely manual approach. We note that any ground-truth (produced either by this approach or through hand-labeling of each instance) will contain impurities in itself on a large dataset, hence the reported quantities capture the accuracy of our method only to a certain extent. However, the low errors (6.56 pixels and 81.2\% IoU on average) approximately quantify the practicality of our method. Our mean IoU metric remains at par with the 80\% IoU reported in LabelFusion while completely eliminating the need for a prior 3D model.

\begin{figure*}
  \includegraphics[width=\textwidth]{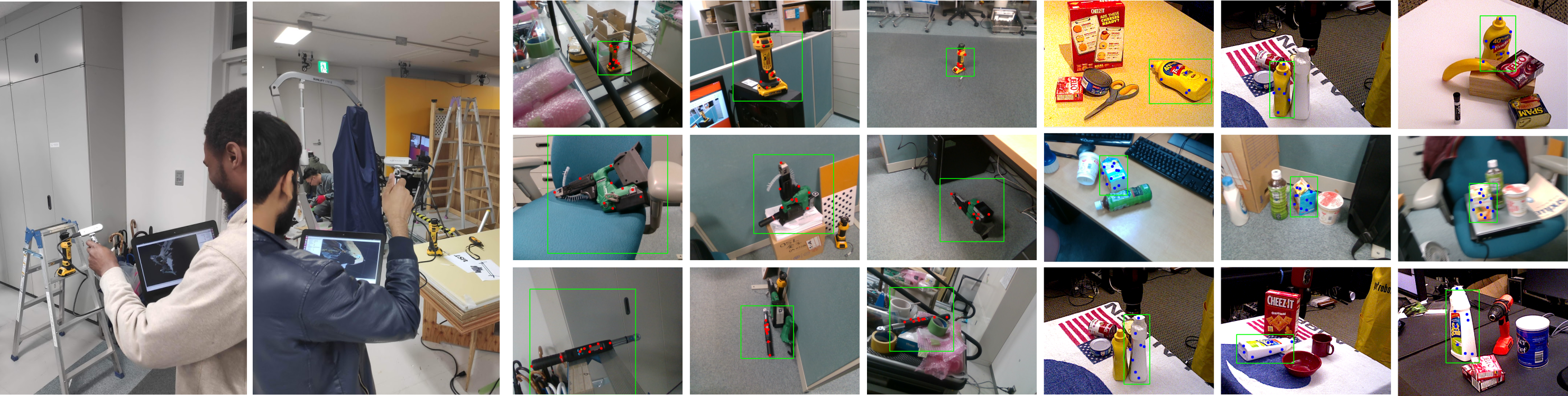}
  \caption{\textbf{Keypoint and bounding-box labels.} Raw real RGB-D data was captured conveniently using a handheld RealSense sensor in 55 scenes for 7 in-house objets with no other hardware requirements. We generated keypoint and bounding-box labels using our approach for all chosen objects (6 are show here) in different scenes without using a prior 3D model from minimal manual annotation. Our approach is easily scalable to a large number of scenes.}
  \label{figure:label_examples}
\end{figure*}

\begin{figure}[t]
  \includegraphics[width=\linewidth]{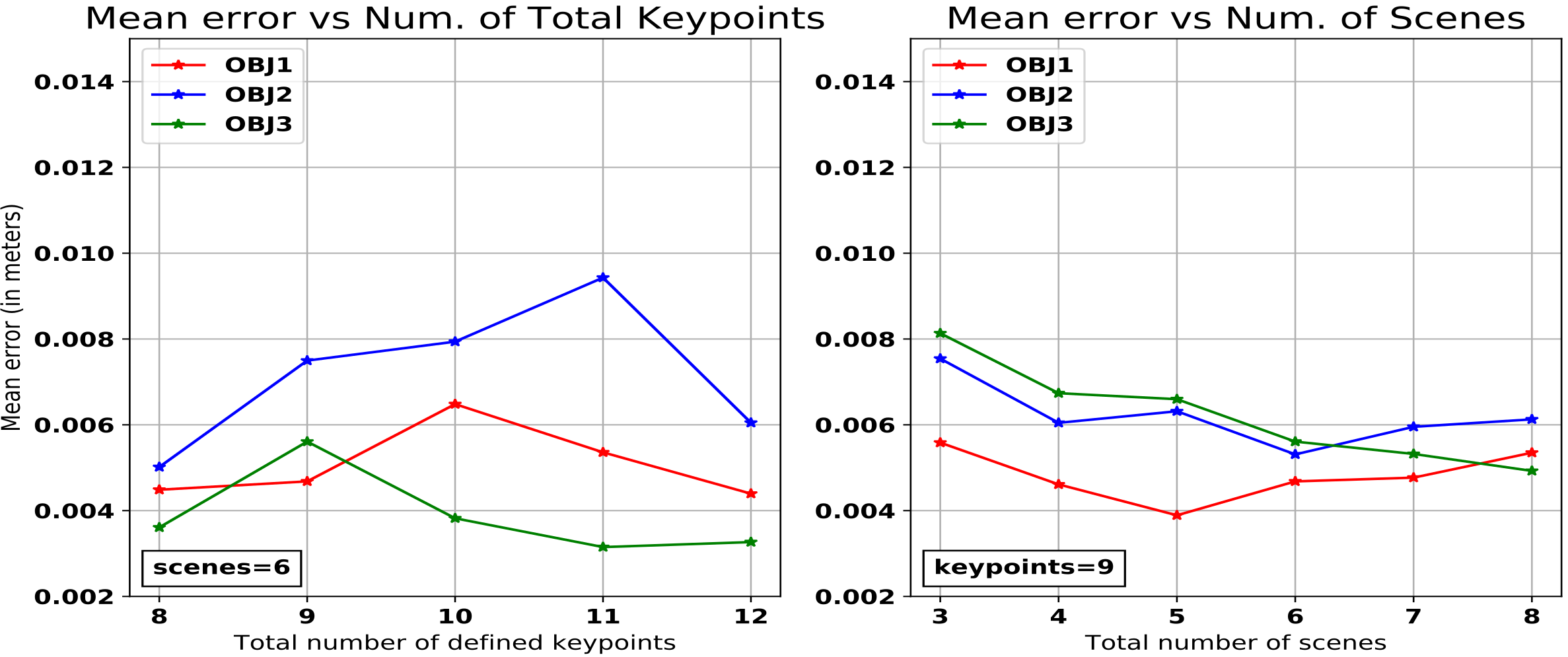}
  \caption{Evaluation of the generated sparse keypoint-based model. We measure the average errors in estimated keypoint position with respect to the total no. of keypoints in (a) and total no. of scenes in (b).}
  \label{figure:error_curves}
\end{figure}

Next, we perform a set of experiments on the following 3 objects: Dewalt Cut-out Tool (OBJ1), Facom Wrench (OBJ2), Hitachi Screw Gun (OBJ3) -- to measure the effect of the number of scenes in $\mathcal{U}$ ($N_s$) and the number of chosen keypoints ($N_k$) on the optimization process. The error in keypoint 3D positions for the 3 objects as a function of the total number of defined keypoints $N_k$ are shown in Figure \ref{figure:error_curves}(a) when $N_s=8$, while Figure \ref{figure:error_curves}(b) shows the mean error as a function of number of scenes $N_s$ when $N_k=9$. As the curves show, the mean positional error remained fairly independent of $N_k$ while the accuracy seemed to improve with $N_s$. This is expected as manual input for the same keypoint in multiple scenes would help remove bias. For $N_k$, a lower number of keypoints makes it harder to cover all faces of the object and a higher number increases the chances of human error in the manual annotation step. In our experience, choosing 6--12 adequately spaced keypoints proved to be sufficient for all objects, while $N_s$ is entirely up to the use-case scenario.

\subsection{Application to 6-DoF Object Pose Estimation}
As we expect the proposed approach to be primarily used for deployment of DL based 6-DoF pose estimation, we evaluate the performance of such a pipeline trained on the labeled dataset generated through our method for OBJ1, OBJ2 and OBJ3. We adopt a keypoint-based pose estimation approach \cite{pavlakos17object3d, rpsingh2020instance}, where a stacked-hourglass network is used for keypoint predictions on RGB images cropped by an object bounding-box detector network (such as SSD \cite{liu2016ssd}, YOLO \cite{redmon2016you:yolo}). Predictions from the hourglass module along with the 3D model correspondences (from $\mathbf{Q}$) are fed through a P\textit{n}P module to obtain the final pose. An overview is depicted in Figure \ref{figure:pose_estimation}.

\begin{figure}[t]
    \centering
    \includegraphics[width=\columnwidth]{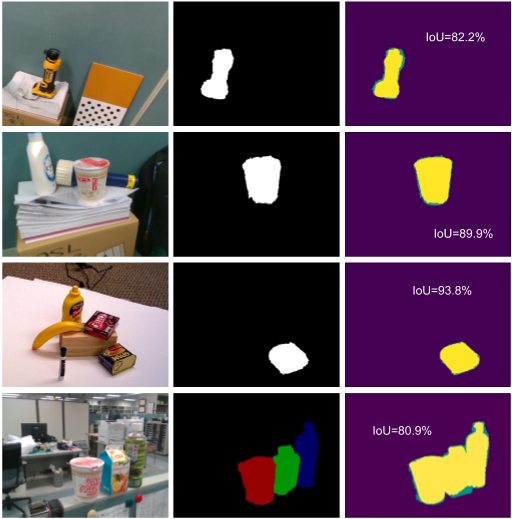}
    \caption{\textbf{Pixel-wise mask labels.} Examples of single and multi-object scenes from the experiments. The generated mask label is shown in the middle while the intersection-over-union computation is on the right.}
    \label{fig:my_label}
\end{figure}

\begin{figure*}[t]
  \includegraphics[width=\textwidth]{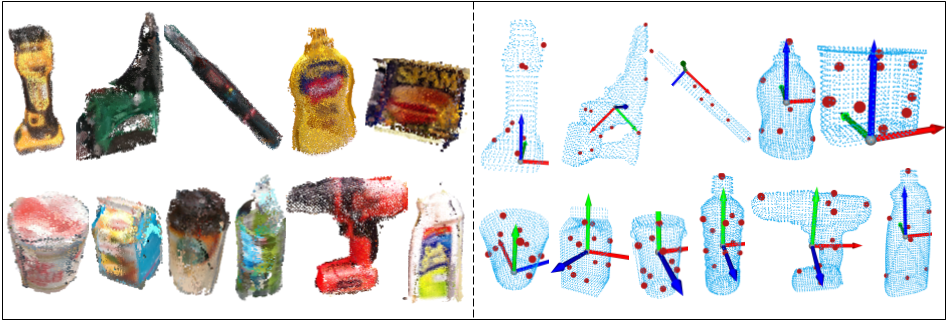}
  \caption{Generated dense models of the experiment objects on the left and the sparse representations overlayed on the CAD models on the right. The dense models are produced by \textit{growing regions} on the scene meshes, seeded by the points of the sparse-representation.}
  \label{figure:sparse_models}
\end{figure*}

\begin{table*}[t]
\caption{Accuracy analysis of generated sparse model and labels.}
\label{tab:table_model_errors}
\begin{center}
\begin{tabular}{c | c | c | c | c | c | c | c}
\hline
& Object & \# KPs & \# scenes & Mean KP Error (3D) & Mean KP Error (2D) & Mean IoU & \# labels\\
& & $(N_{k})$ & $(N_{s}=$len$(\mathcal{U}))$ & (in \textit{mm}) & (in \textit{pixels}) & & (sampled@3Hz)\\
\hline
01 & \textit{Dewalt Cut-out Tool} & 8 & 11 & 1.00 & 4.23 & \textbf{81.09} & 3043\\
\hline
02 & \textit{Facom Electronic Wrench} & 6 & 10 & 2.69 & 6.66 & 71.54 & 2188\\
\hline
03 & \textit{Hitachi Screw Gun} & 9 & 9 & 5.29 & 6.08 & 75.73 & 2172\\
\hline
04 & \textit{Cup Noodle} & 10 & 6 & 0.5 & 3.97 & \textbf{88.95} & 860\\
\hline
05 & \textit{Lipton} & 11 & 6 & 2.1 & 8.02 & \textbf{83.19} & 1029\\
\hline
06 & \textit{Mt. Rainier Coffee} & 11 & 8 & 1.6 & 7.40 & \textbf{82.49} & 1036\\
\hline
07 & \textit{Oolong Tea} & 11 & 5 & 1.1 & 3.65 & \textbf{83.6 }& 850\\
\hline
08 & \textit{Mustard Bottle} & 7 & 5 & 1.17 & 7.15 & \textbf{83.29} & 947\\
\hline
09 & \textit{Potted Meat Can} & 10 & 8 & 1.02 & 8.83 &\textbf{ 83.49} & 1391\\
\hline
10 & \textit{Bleach Cleanser} & 8 & 9 & 3.37 & 9.73 & \textbf{81.5} & 1525\\
\hline
11 & \textit{Power Drill} & 12 & 9 & 1.52 & 6.43 & 78.25 & 1346\\
\hline
 & \textbf{Mean} & - & - & \textbf{1.94} & \textbf{6.56} & \textbf{81.2} & -\\
\hline
\end{tabular}
\end{center}
\end{table*}

We trained the YOLO and the stacked-hourglass networks for each of the 3 objects from scratch on images sampled from the RGB-D videos of all scenes at 3Hz with a 90-10 split for training and testing respectively. The YOLO network was trained on the default hyperparameters while the stacked-hourglass network was trained on mostly the same hyperparameters as \cite{pavlakos17object3d} (though we implemented the keypoint detector in PyTorch instead of Lua originally). After the keypoint detection stage, we computed the 6-DoF object pose using OpenCV's E-P\textit{n}P method, using the generated sparse model $\mathbf{Q}$ in the test pipeline and the hand-modelled 3D CAD for the benchmark pipeline. For benchmarking, we trained the networks again from scratch on the same set of raw images but using the ground-truth labels and compared the errors in estimated 6-DoF poses. Rotation error was computed using the following geodesic distance formula: \par

\begin{equation}
\begin{gathered}
\Delta(R_1, R_2)
= \frac{{\lVert log(R_1^T R_2) \rVert}_{F}}{\sqrt{2}}.
\end{gathered}
\end{equation}

The median errors in rotation and translation for the 3 objects for the pipeline, trained on the generated dataset (OURS) and ground-truth dataset (MANUAL), are consolidated in Table \ref{tab:table_pose_errors}. The results indicate that our proposed approach, while reducing the human time required for the entire procedure by several orders of magnitude, can be used to generate labeled datasets for training pose estimation pipelines to remarkable accuracy (comparable to that trained on manually generated dataset).

\subsection{Application to Object Pixel-wise Segmentation}
We investigate the suitability of the generated mask labels for the training of existing object segmentation methods. We selected an open-source implementation of Mask R-CNN \cite{matterport_maskrcnn_2017} -- a popular approach for pixel-wise segmentation of objects in RGB images.

From the labeled dataset generated in our experiments, we selected a subset of images (sampled at 3Hz) and again created a 90/10 train-test split. Although our tool is capable of generating labels for multi-object scenes, here we limit our analysis to single-object scenes for the sake of simplicity. We trained the model for 11 classes of objects (refer to Table \ref{tab:table_model_errors} for names), with default initialization of the weights. The ResNet101 backbone was chosen for the architecture and training images were square cropped to $512\times512$ size. The training was performed on ``head" layers for the first 20 epochs followed by ``all" layers for then next 20. Other hyperparameters were set to default.

\begin{figure}[t]
  \includegraphics[width=\columnwidth]{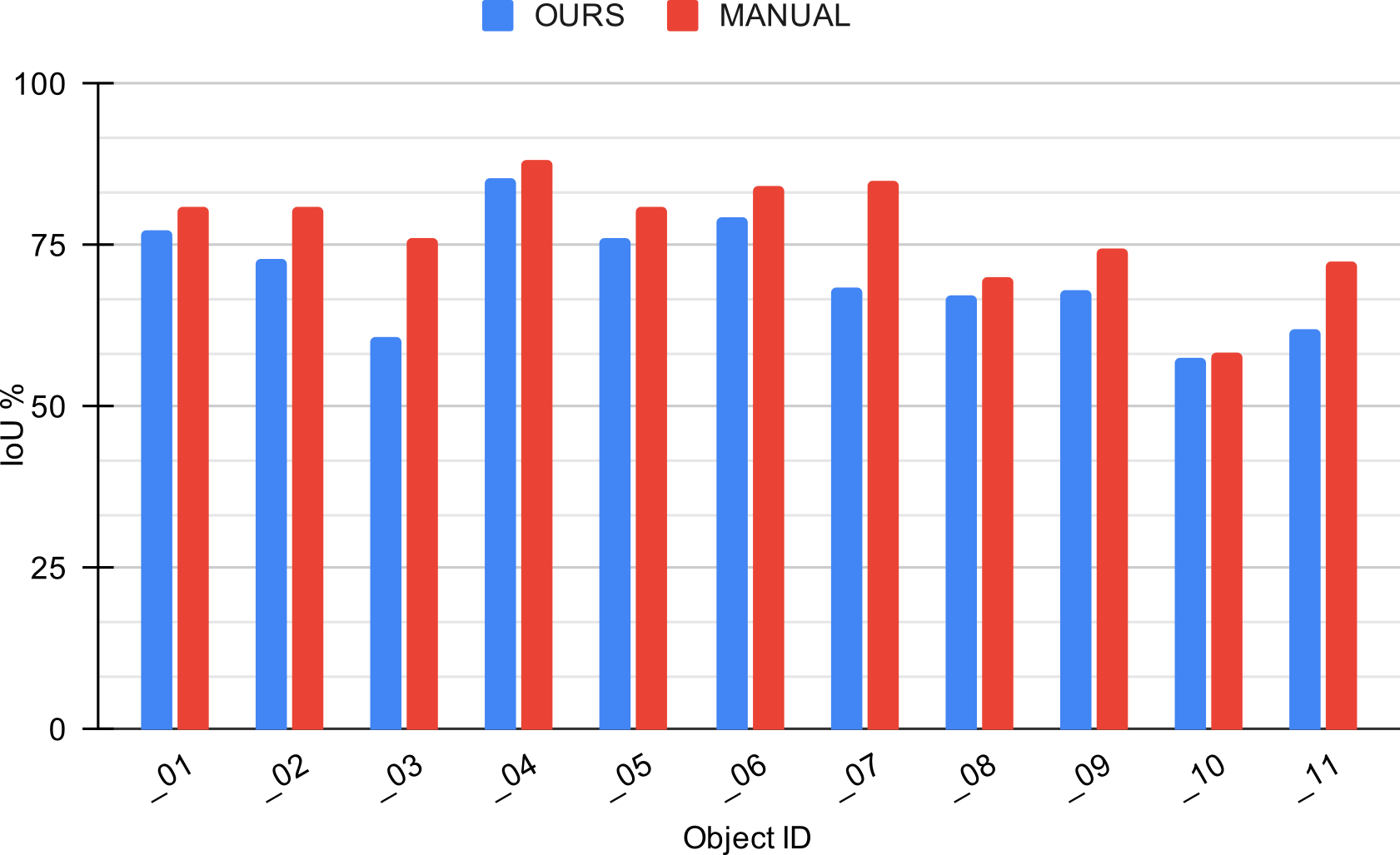}
  \caption{\textbf{Performance of Mask R-CNN trained on our dataset.} Mean IoU scores for each of the 11 objects in the dataset.}
  \label{figure:segmentation}
\end{figure}

\addtolength{\textheight}{-2cm}   
                                  
The purpose here was to measure the performance of the segmentation network trained on the automatically generated dataset (OURS) and compare it to the segmentation network trained on the ground-truth dataset (MANUAL). In both cases, the trained network was evaluated against the test subset of the ground-truth dataset. The Intersection-over-Union metric was measured and reported in Figure \ref{figure:segmentation}. As the object-wise IoU metric remains comparable in both cases, we argue that training on our dataset (which takes a lot less manual effort to generate) provides similar performance for real applications.
\begin{table}[t]
\caption{Errors (Median) in Object Pose Estimation}
\label{tab:table_pose_errors}
\begin{center}
\begin{tabular}{c | c | c | c | c}
\hline
& \multicolumn{2}{c|}{Position} & \multicolumn{2}{c}{Orientation}\\
& \multicolumn{2}{c|}{(in \textit{mm})} & \multicolumn{2}{c}{(in \textit{degrees})}\\
\hline
& OURS & MANUAL & OURS & MANUAL\\
\hline
\textit{OBJ1} & 9.93 & \textbf{9.23} & 0.78 & \textbf{0.69}\\
\hline
\textit{OBJ2} & \textbf{3.66} & 4.75 & 1.35 & \textbf{1.23}\\
\hline
\textit{OBJ3} & \textbf{6.77} & 7.52 & \textbf{4.46}     & 4.80\\
\hline
\end{tabular}
\end{center}
\end{table}

\section{CONCLUSION \& FUTURE WORK}

Through this paper, we have presented a technique for rapidly generating large datasets of labeled RGB images that can be used for training of deep CNNs for various applications. Our pipeline is highly automated --- we gather input from the human user for a few keypoints in only one RGB per scene. We do not require any complicated hardware setups like robot manipulators and turntables or sophisticated calibration procedures. In fact, the only hardware requirements are -- a calibrated RGB-D sensor and the object itself -- consequently, making it easier for the user to acquire the dataset in different environments.

We validated the effectiveness of the proposed method by using it to rapidly produce more than 150,000 labeled RGB images for 11 objects and subsequently, using the dataset to train a pose estimation pipeline and a segmentation network. We evaluated the accuracy of the trained networks to establish practicality and applicability of the labeled datasets as a solution to 6-DoF pose for real-world robotic grasping and manipulation tasks.

Future work in this direction would focus on making the sparse model of an object to be used as a canonical representation of the object's class to help class-specific pose estimation. Also, improving the quality of the segmented dense models by the means of a better algorithm would help improve the quality of labels.


\bibliographystyle{IEEEtran}
\bibliography{IEEEabrv,bibliography.bib}
\end{document}